%% file: main.tex
\documentclass{article}

\usepackage{arxiv}%
\usepackage{graphicx}
\usepackage{hyperref}

\usepackage{algorithm}
\usepackage{algorithmic}
\usepackage{rotating}
\usepackage{amsmath}
\usepackage{caption}
\usepackage{subcaption}
\usepackage{longtable}
\usepackage{booktabs}
\usepackage{multirow}
\usepackage{makecell}
\usepackage{natbib}
\usepackage{dsfont}
\usepackage{tabularx}
\usepackage{url}
\newcolumntype{C}[1]{>{\centering\arraybackslash}p{#1}}

\begin{document}
\title{A Federated Cox Model with Non-Proportional Hazards}

\author{ {Dekai Zhang} \\
	Department of Computing\\
	Imperial College London\\
	London, SW7 2BX \\
	\texttt{dz819@imperial.ac.uk} \\
	%% examples of more authors
	\And
	{Francesca Toni} \\
    Department of Computing\\
	Imperial College London\\
	London, SW7 2BX \\
	\texttt{f.toni@imperial.ac.uk} \\
	\And
	{Matthew Williams} \\
    Department of Surgery \& Cancer\\
	Imperial College London\\
	London, SW7 2BX \\
	\texttt{matthew.williams@imperial.ac.uk} \\
	%% \AND
	%% Coauthor \\
	%% Affiliation \\
	%% Address \\
	%% \texttt{email} \\
	%% \And
	%% Coauthor \\
	%% Affiliation \\
	%% Address \\
	%% \texttt{email} \\
	%% \And
	%% Coauthor \\
	%% Affiliation \\
	%% Address \\
	%% \texttt{email} \\
}

\maketitle              % typeset the header of the contribution
\input{01abstract}
\input{02intro}
\input{03background}
\input{04model}

\input{05experiments}

\input{06discussion}

\subsubsection*{Acknowledgements.}
This work was supported by the UKRI CDT in AI for Healthcare \url{http://ai4health.io} (Grant No. EP/S023283/1)
\appendix
\input{appxKM}

%
% ---- Bibliography ----
%
% BibTeX users should specify bibliography style 'splncs04'.
% References will then be sorted and formatted in the correct style.
%
\bibliographystyle{unsrtnat}
\bibliography{references}

\end{document}

%% file: 01abstract.tex
\begin{abstract}
Recent research has shown the potential for neural networks to improve upon classical survival models such as the Cox model, which is widely used in clinical practice. Neural networks, however, typically rely on data that are centrally available, whereas healthcare data are frequently held in secure silos. We present a federated Cox model that accommodates this data setting and also relaxes the proportional hazards assumption, allowing time-varying covariate effects. In this latter respect, our model does not require explicit specification of the time-varying effects, reducing upfront organisational costs compared to previous works. We experiment with publicly available clinical datasets and demonstrate that the federated model is able to perform as well as a standard model. 

% \keywords{Survival analysis \and Federated learning \and Non-proportional hazards}
\end{abstract}

%% file: 02intro.tex
\section{Introduction}
\label{sec:intro}

Estimating how long patients might live for is a key task in clinical medicine, and is a common question from patients. Survival analysis is the statistical branch used to perform these estimates, which can range in its application from predicting death following diagnosis to loan defaults or machine part failures. Amongst survival models, the Cox model \citep{Cox1972} is one of the most widely used. 

Machine learning techniques have received attention for their potential to improve upon the performance of the Cox model. Many recent efforts \citep{Katzman2018DeepSurv:Network,Luck2017DeepAnalysis,Kvamme2019Time-to-eventRegression} have exploited neural networks (NNs) to model more complex relationships as well as enable typically unsupported input data types such as images \citep{Zhu2017DeepImages,Zhu2017WSISA:Images,Li2019,Bello2019Deep-learningPrediction}. Notwithstanding, the Cox model has remained the standard in survival analysis \citep{Wang2019MachineSurvey}. Indeed, the adoption of machine learning has progressed haltingly in many areas of healthcare \citep{Kelly2019KeyIntelligence}.

One challenge lies in the distributed nature of healthcare data \citep{Wiens2019}. In much of machine learning, data are centralised, whereas privacy concerns often result in secure data ``silos'' in healthcare. Federated learning (FL) accommodates this decentralised data environment and has shown promise in clinical contexts \citep{Rieke2020}.

Despite a fast emerging literature in FL, there has been scant work on federated survival analysis. \citet{Andreux2020FederatedModels} propose a federated Cox model that is closest to this paper. The standard Cox model is, however, limited in that it can only correctly model proportional hazards. We take an alternative approach allowing us to embed time-varying covariate effects (\emph{non}-proportional hazards) directly in the architecture, potentially reducing organisational setup costs for federations. Such effects are relevant to adapt models for patients such as those with breast cancer where the proportional hazards assumption has been shown to be violated \citep{Bellera2010VariablesCancer,Coradini2000Time-dependentCancer,Gore1984RegressionAuthor}.

In the following, we briefly discuss relevant background and highlight related work (Section~\ref{sec:background}), before defining our model (Section~\ref{sec:model}), instantiating it with different hazards assumptions and presenting our experiments with real-world clinical datasets (Section~\ref{sec:experiments}). Section~\ref{sec:discussion} concludes with potential directions for future work.

%% file: 03background.tex
\section{Background and Related Work}
\label{sec:background}
The promise of greater control over data ownership and enhanced privacy that FL affords has generated interest in the healthcare community. Few works, however, have investigated the intersection between survival analysis and FL. We present background on each of these areas separately before discussing their intersection relevant for this work. 

\subsubsection{Background on Federated Learning (FL).}
\label{sec:background:fl}
FL is a framework for decentralised data that cannot be shared due to their sensitive content or prohibitive communication costs \citep{BrendanMcMahan2017}. In the context of healthcare, patient data may be kept in this way by the clinical unit (e.g., the hospital) at which the patient was treated. In the following, we will simply refer to these data-keeping units as federation members or centres. Typically (and in this paper), the federated objective is to minimise $\mathcal{L}_F(X,\phi)$ with respect to $\phi$ with:

\begin{equation}
    \mathcal{L}_F(X,\phi) \!=\! \sum_{k \in K} w_k \mathcal{L}_k(X_k,\phi)
    \label{eq:fl-objective}
\end{equation}
\noindent
where $\mathcal{L}_F$ represents the global loss: an average of the local losses $\mathcal{L}_k$ computed by the federation members in $K$ on their own data $X_k$ weighted by $w_k$, where $\phi$ represents the model parameters. Typically, each member customises $\phi$ for a number of local optimisation rounds before aggregating the customised $\phi$ for a new global consensus model. 

\subsubsection{Background on Survival Analysis.}
\label{sec:background:survival}
Survival analysis estimates the time to an event for a population $N$ with data $\mathcal{D} = \{(x_i, t_i, s_i)\}_{i \in N}$ where each person $i$ has covariates $x_i = (x_{i1}, ..., x_{ip})^\top$, a time of observation $t_i$ and an indicator $s_i \in \{0,1\}$ which equals 1 if $i$ has experienced the event or 0 if not, i.e., if $i$ is \emph{censored}.

% A distinguishing feature is its ability to handle \emph{censored} data. By way of illustration, a researcher may wish to study a population's mortality and therefore collect the ages at death. For part of the population instead of the age at death only the age at the time of data collection will be known. Such a population is said to be \emph{right-censored}.\footnote{For a comprehensive discussion of other censoring issues in survival analysis, see \citep{Leung1997}.} The population $N$ can be described by the set of tuples:

The \emph{Cox model} \citep{Cox1972} is one of the most widely used survival models. It defines a hazard function $h$, which expresses the rate of failure at time $t$ subject to survival until then as follows:

\begin{equation}
    \begin{split}
    h(t| x_i) &= P(T = t | T > t-1) \\
                &= h_0(t) exp[g(x_i)] \text{ with } g(x_i) = \beta^\top x_i
    \end{split}
    \label{eq:basic_cox}
\end{equation}
\noindent
where $h_0(t)$ is some baseline hazard and where $\beta = (\beta_1,...,\beta_p)^\top$ is a coefficient vector. Later works replace the linear predictor $\beta^\top x_i$ with NNs $g_\phi(x_i)$, demonstrating competitive performance \citep{Faraggi1995AData,Katzman2018DeepSurv:Network}. The coefficients are estimated by minimising the negative partial log-likelihood given by: 

\begin{equation}
    -\sum_{i \in N} s_i [ g(x_i) - \log(\sum_{j \in R_i} exp[g(x_j)])   ]
    \label{eq:cox_llh}
\end{equation}

% \begin{equation}
%     -\sum_{i \in N} s_i \left[ g(x_i) - \log\left(\sum_{j \in R_i} exp[g(x_j)]\right)   \right]
%     \label{eq:cox_llh}
% \end{equation}
\noindent
where $R_i = \{j \in N:t_j \geq t_i\}$ denotes the individuals who are still \emph{at risk} when $i$ experiences the event. 

In a federated setting, this loss generally cannot be decomposed into local losses due to the logarithmic term, as the risk set $R_i$ can contain individuals from centres other than the one of $i$. This therefore does not match the formulation of Eq.~\ref{eq:fl-objective}. The hazard function also assumes proportional hazards (PH) -- differences in covariates result in constant proportional differences in hazards. Over long time horizons, this can be restrictive \citep{Kvamme2019Time-to-eventRegression,Bellera2010VariablesCancer,Antolini2005AData}.

\subsubsection{State-of-the-Art in Federated Survival Analysis.}
\label{sec:background:fedsurv}
Our work is situated in the intersection of federated learning and survival analysis and proposes a novel framework. A handful of works have already proposed such frameworks of which we provide a brief overview here. The works of \citet{Lu2015WebDISCO:Sharing} and \citet{Dai2020VERTICOX:Multipliers} embody one approach which relies on substantial sharing of summary statistics over the local datasets in every training iteration. This differs in spirit from FL where more abstract parameters are shared and, often, infrequently so. Moreover, their models are based on linear predictors and do not address integration with NNs. 

Recent work by \citet{Andreux2020FederatedModels} is closest to our approach. Their model exploits a discretisation of the Cox model (also by \citet{Cox1972}) with an NN-based predictor:

\begin{equation}
    \dfrac{h(t|x)}{1-h(t|x)} = \dfrac{h_0(t)}{1-h_0(t)}exp[g_\phi(x_i)] 
    \label{eq:discrete_cox_nn}
\end{equation}
\noindent
which can be rewritten in a sigmoid form:

\begin{equation}
    h(t|x_i) = \dfrac{1}{1 + exp[-(\alpha_t + g_\phi(x_i))]}
    \label{eq:discrete_cox_nn_sigmoid}
\end{equation}
\noindent
where $\alpha_t = \log(\frac{h_0(t)}{1-h_0(t)})$. 

They follow \citet{Craig2021SurvivalProblem} in estimating this function like a logistic regression with negative log-likelihood:
% \begin{equation}
%     \begin{split}
%     \ell(X) = - \sum_{i \in N} \sum_{k=1}^{t_i} & \Big[ y_{ik} \log[h(k|x_i)] \\ &
%                     + (1-y_{ik}) \log[(1-h(k|x_i))] \Big]
%     \end{split}
%     \label{eq:discrete_llh}
% \end{equation}

\begin{equation}
    % \begin{split}
    - \sum_{i \in N} \sum_{k=1}^{t_i} [ y_{ik} \log[h(k|x_i)] + (1-y_{ik}) \log[(1-h(k|x_i))] ]
    % \end{split}
    \label{eq:discrete_llh}
\end{equation}
\noindent
where $y_{ij} = \mathds{1}\{t_j = t_i, s_i = 1\}$. Importantly, this loss does not depend on risk sets and is therefore separable -- each centre's loss only depends on local data -- recovering the federated objective (Eq.~\ref{eq:fl-objective}).
 
\citet{Andreux2020FederatedModels} demonstrate that a federation of this model can draw even in performance with a model trained on pooled data. This is, however, only shown with aggregation after every local optimisation round -- a setup that may need to differ in practice \citep{Rieke2020} -- and assuming PH, as their predictor $g_\phi(x_i)$ is time-invariant. 

We note that non-PH can be admitted to their model by including time interactions (giving $g_\phi(x_i,f(t))$) -- an approach sometimes taken in standard Cox models -- as demonstrated on pooled data by \citet{Craig2021SurvivalProblem}. This could, however, introduce a dependency on the specification of $f(t)$ and its interactions.  Crucially, this may add to the organisational setup costs of a federation: even though interactions could be learned, $f(t)$ needs to be fully specified and agreed upon in advance. In contrast, we follow \citet{Gensheimer2019ANetworks} by making the choice between PH and non-PH a binary decision over the architecture of the output layer. 

%% file: 04model.tex
\section{Model}
\label{sec:model}
We build upon the discretised Cox model and detail how the PH assumption is relaxed and formulate the federated objective. We describe a discretisation procedure and an optional interpolation scheme for smooth predictions. Lastly, we outline two complementary performance metrics.
% \FT{ADD A SHORT NARRATIVE TO INTRODUCE THE VARIOUS SUBSECTIONS...WHAT PROBLEM ARE YOU TRYING TO SOLVE? ETC}

\paragraph{Non-Proportional Hazards.}
We use a discretised Cox model (Eq.~\ref{eq:discrete_cox_nn_sigmoid}) but parameterised with a time-varying, NN-based predictor $g_{\phi,t}(x)$. Following \citet{Gensheimer2019ANetworks}, we allow for non-PH by fully connecting the output layer to the previous layer. The output layer thus encodes time-varying covariate effects in time-specific weights. A sigmoid is used to retrieve the hazard rates.

\begin{figure}[h]
    \centering
    \begin{subfigure}[b]{0.43\linewidth}
        \centering
        \includegraphics[width=\linewidth]{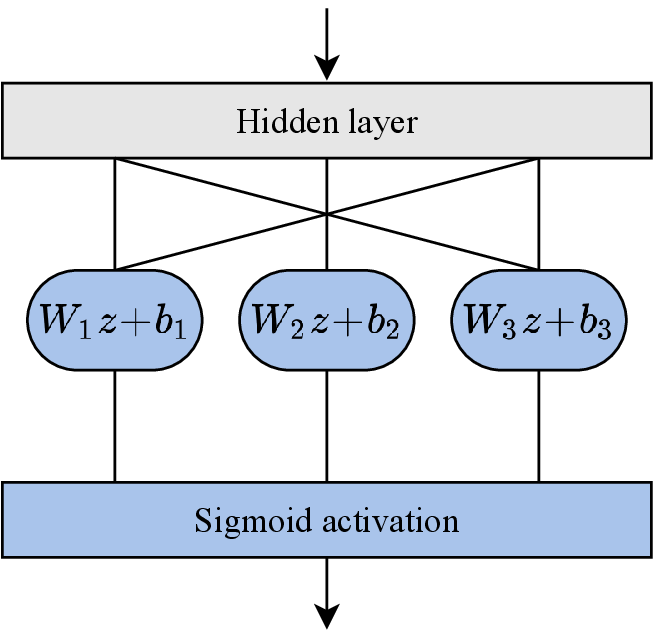}%
        \caption{Non-PH}
        \label{fig:nn_detail_nph}
    \end{subfigure}
    \hfill
    \begin{subfigure}[b]{0.43\linewidth}
        \includegraphics[width=\linewidth]{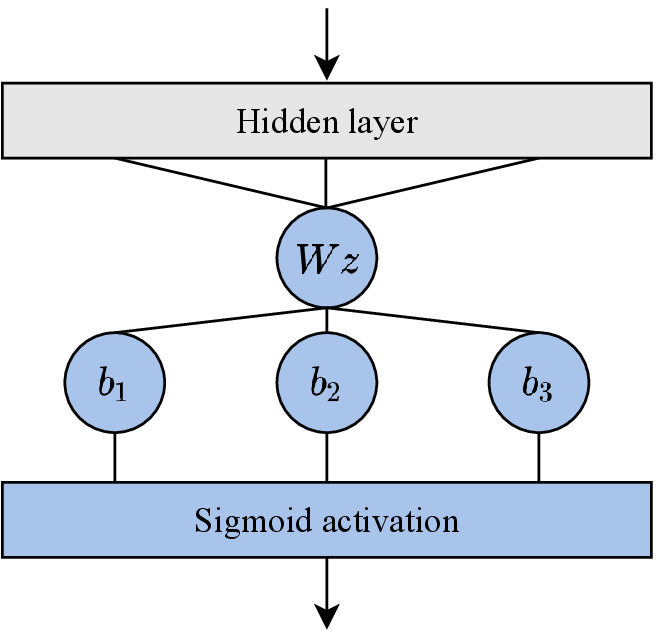}
        \caption{PH}
        \label{fig:nn_detail_ph}
    \end{subfigure}
    % \qquad
    \caption{Output components for 3 time steps.}
    \label{fig:nn_detail}
\end{figure}

For PH, the output component is split into a first layer with a single neuron and no bias. The output of this neuron is passed into a second layer with as many neurons as time steps. This captures time-varying baseline hazards in the second layer and time-invariant covariate effects in the first. The difference in components is illustrated in Figure~\ref{fig:nn_detail}.

\paragraph{Federated Objective.} 
To conform to a federated formulation (Eq.~\ref{eq:fl-objective}), we split the objective (Eq.~\ref{eq:discrete_llh}): 

\begin{equation}
    \begin{split}
    \mathcal{L}_{F} &= \sum_{k \in K} w_k \mathcal{L}_k(X_k,\phi) \\
    &= -\sum_{k \in K} \frac{|N_k|}{|N|} \sum_{i \in N_k} \sum_{j=1}^{t_i} \Big[y_{ij} \log[h(j|x_i)] + (1-y_{ij})  \log[(1-h(j|x_i))]\Big]
    \end{split}
    \label{eq:fed_loghaz_objective}
\end{equation}

\noindent
where $y_{ij} = \mathds{1}\{t_j = t_i, s_i = 1\}$. Each centre calculates $\mathcal{L}_k(X_k, \phi)$ on its own subset of the population $N_k$. We adapt the FedAvg algorithm of \citet{BrendanMcMahan2017} to minimise this loss (Algorithm~\ref{alg:fedavg}).

\begin{algorithm}[htbp]
\caption{Procedure to optimise the federated objective.}
\label{alg:fedavg}
% \textbf{Input}: Your algorithm's input\\
% \textbf{Parameter}: Optional list of parameters\\
% \textbf{Output}: Your algorithm's output
\begin{algorithmic}[1] %[1] enables line numbers
\STATE Initialise global model with $\phi_0$
\FOR{each round $t = 1, ..., T$} 
    \FOR{each centre $k = 1, ..., K$ in parallel}
        \STATE Send $\phi_{t-1}$ to centre k
            \FOR{$ \text{each local round } b = 1, ..., B$}
                \STATE Local update $\phi_t^k \leftarrow \phi_t^k - \lambda \nabla \mathcal{L}_k(X_k, \phi_t^k)$
            \ENDFOR
        \STATE Receive $\phi_t^k $ from centre k
    \ENDFOR
    \STATE Aggregate $\phi_t \leftarrow \sum_{k\in K} \frac{|N_k|}{|N|} \phi_t^k$
\ENDFOR
\RETURN $\phi_T$
% \WHILE{condition}
% \STATE Do some action.
% \IF {conditional}
% \STATE Perform task A.
% \ELSE
% \STATE Perform task B.
% \ENDIF
% \ENDWHILE
\end{algorithmic}
\end{algorithm}

\paragraph{Discretisation.}
The model operates on discretised time, so that $t$ indexes into a set of intervals $[\tau_{t-1}, \tau_t)$. Following \citet{Kvamme2019ContinuousNetworks} we discretise on Kaplan-Meier quantiles. Defining the survival curve $S(\tau) = S(\tau-1)(1 - h(\tau))$, the quantiles $\{\tau_1, \tau_2, ..., \tau_m\}$ can be obtained as:

\begin{equation}
    S(T = \tau_j) - S(T = \tau_{j+1}) = \dfrac{1 - S(T = \tau_{max})}{m}
\label{eq:KM_discretisation}
\end{equation}
\noindent
for $j = \{0, 1, ..., m-1\}$. This discretisation procedure yields a set of steps $\{\tau_1, \tau_2, ..., \tau_m\}$ where each step results in the same decrease in survival (an illustration is provided in Appendix~\ref{sec:appx}).

\paragraph{Interpolation.}
To smooth step-wise predictions, we use constant density interpolation \citep{Kvamme2019ContinuousNetworks}. Letting $\widetilde{S}(\tau)$ denote the interpolation of the survival curve $S(\tau)$, we then have:

\begin{equation}
    \widetilde{S}(\tau) \!=\! S(\tau_{j-1}) + \left[S(\tau_j) - S(\tau_{j-1})\right] \dfrac{\tau - \tau_{j-1}}{\tau_j - \tau_{j-1}}
\label{eq:interpolation_cdi}
\end{equation}

\noindent
for a given time $\tau \in (\tau_{j-1}, \tau_j]$. Intuitively, the step survival curve is linearly interpolated between any adjacent steps, resulting in constant densities in the corresponding interval (an illustration is provided in Appendix~\ref{sec:appx}).

\paragraph{Performance Metrics -- Concordance.}
We use the time-dependent concordance index \citep{Antolini2005AData}, or c-index, which is a discriminative measure for how well the model ranks the relative survival between patient pairs, expressed as:

\begin{equation}
    P(S(t_i|x_i) < S(t_i|x_j) \ \& \ t_i < t_j \ \& \ s_i = 1) 
    \label{eq:concordance_td}
\end{equation}

\noindent
which is estimated as follows: 

\begin{equation}
    \widehat{c} = \dfrac{\sum_{i \in N} \sum_{j \in N, j\neq i} conc_{ij}}
        {\sum_{i \in N} \sum_{j \in N, j\neq i} comp_{ij}}
    \label{eq:c-index_td_estimated}
\end{equation}

\begin{equation}
    comp_{ij} = \mathds{1}\{t_i < t_j \ \& \ s_i = 1\} + \mathds{1}\{t_i = t_j \ \& \ s_i = 1 \ \& \ s_j=0\}
    \label{eq:c-index_td_estimated_comp}
\end{equation}

\begin{equation}
    conc_{ij} = \mathds{1}\{S(t_i|x_i) < S(t_i|x_j)\}\ comp_{ij}
    \label{eq:c-index_td_estimated_conc}
\end{equation}

\paragraph{Performance Metrics -- Calibration.}
While the c-index measures the discriminative performance of the model, it does not measure how well \emph{calibrated} these estimates are (an illustration is provided in Appendix~\ref{sec:appx}).

As a measure of calibration, we follow \citet{Graf1999AssessmentData} who propose a Brier score for use with censored data defined as follows:

\begin{equation}
    BS(t) = \dfrac{1}{|N|}\sum_{i\in N}w_i(t)\Big(y_i(t) - h(t|x_i)\Big)^2 
    \label{eq:graf_brier}
\end{equation}

\begin{equation}
    w_i(t) = \begin{cases}
    s_i / G(t), & \text{if $t_i \leq t$}\\
    1 / G(t), & \text{if $t_i > t$}
    \end{cases}
    \label{eq:graf_brier_weights}
\end{equation}

\noindent
where $y_{i}(t) = \mathds{1}\{t_i = t\}$ and $G(t)$ is the Kaplan-Meier estimate of the censoring distribution (i.e., estimated on $\{(x_i, t_i, 1-s_i)\}_{i\in N}$). To measure calibration across the entire time horizon, we numerically integrate the Brier score using 100 time points \citep{Kvamme2019Time-to-eventRegression}.

%% file: 05experiments.tex
\section{Experiments}
\label{sec:experiments}
We introduce the datasets we experiment on, describe the setup of a simulated federation and instantiate our model with three different linearity and hazards assumptions, and present our results.\footnote{For source code, see \url{https://github.com/dkaizhang/federated-survival}.} 

\subsection{Datasets}
\label{sec:experiments:data}

\begin{table}[h]
\centering
  \begin{tabularx}{\columnwidth}{l X X X X X}
  Dataset & Size & Features & Prop. censored & Last event\\
  \toprule
  METABRIC & 1,904 & 9 & 42\% & 355 days\\
  SUPPORT & 8,873 & 14  & 32\% & 1,944 days\\
  GBSG & 2,232 & 7  & 43\% & 83 days\\
  \bottomrule
  \end{tabularx}
  \caption{Overview of datasets.}
  \label{table:data}
\end{table}

We experiment on three clinical datasets (Table~\ref{table:data}; for Kaplan-Meier curves see Appendix~\ref{sec:appx}) made available by \citet{Katzman2018DeepSurv:Network}, namely the Molecular Taxonomy of Breast Cancer International Consortium (METABRIC), the Study to Understand Prognoses Preferences Outcomes and Risks of Treatment (SUPPORT), and the Rotterdam tumour bank and German Breast Cancer Study Group (GBSG). METABRIC and GBSG both relate to breast cancer patients, a group for whom non-PH have been noted \citep{Bellera2010VariablesCancer,Coradini2000Time-dependentCancer}, while SUPPORT presents serious hospitalisations for a second application area.

\subsection{Setup}
\label{sec:experiments:setup}
\begin{figure}[h]
    \centering
    \begin{subfigure}[b]{0.4\linewidth}
        \centering
        \includegraphics[width=\linewidth]{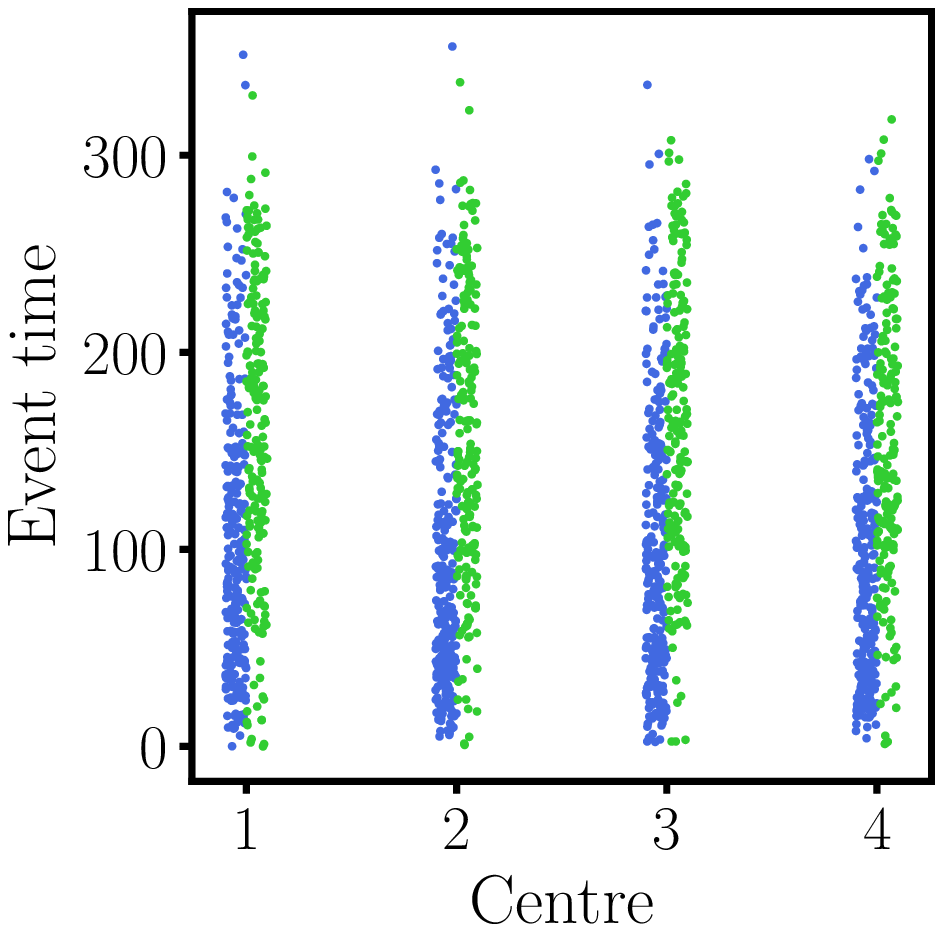}%
        \caption{IID}
        \label{fig:iid}
    \end{subfigure}
    \hfill
    \begin{subfigure}[b]{0.4\linewidth}
        \includegraphics[width=\linewidth]{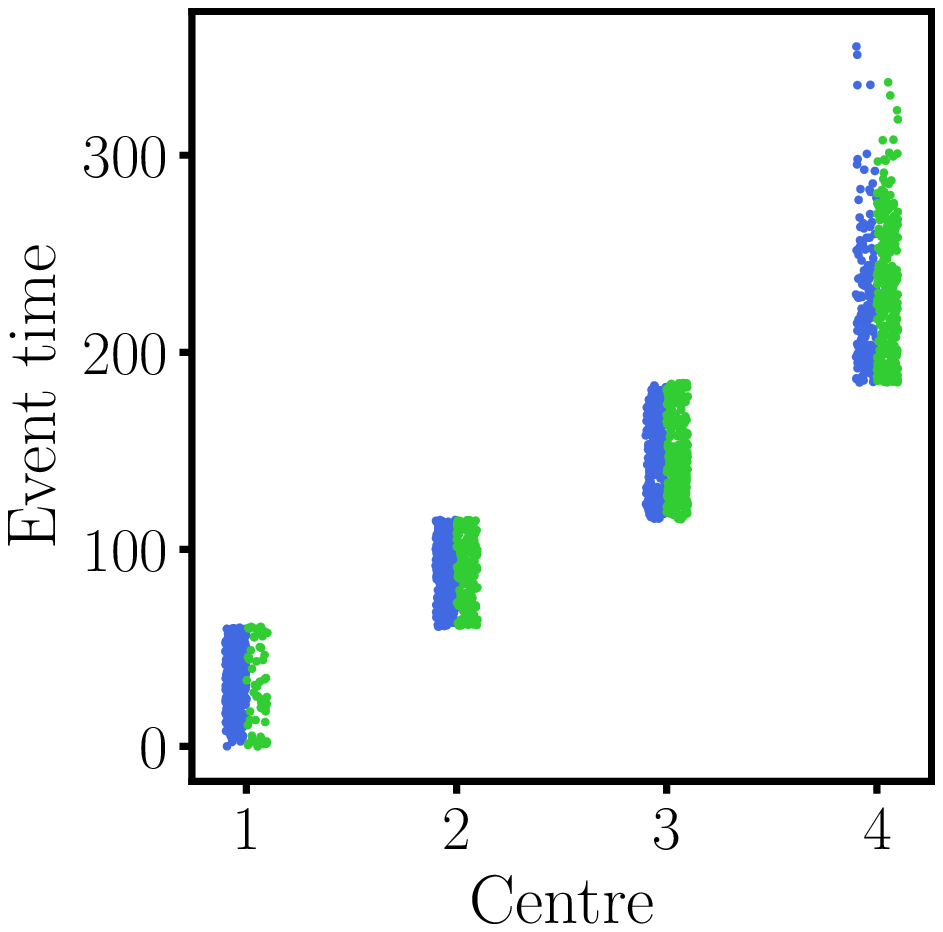}
        \caption{Non-IID}
        \label{fig:noniid}
    \end{subfigure}
    % \qquad
    \caption{Event time distribution for IID and non-IID data using stratification by event time on METABRIC.}
    \label{fig:stratification}
\end{figure}

We simulate two federated data cases: In the first, data are randomly distributed (``IID''), simulating the case of each centre seeing a similar sample of the patient population. In the second, data are stratified on the time to event (``Non-IID''), simulating the case that each centre sees a non-overlapping quantile of the population -- from centre 1 seeing only the shortest survivals leading to centre 4 with the longest survivals (Figure~\ref{fig:stratification}). For comparability, we maintain the total number of local training rounds at 100. A pooled data baseline is provided (``Pooled'' -- no distinction between local and global rounds). In all cases, 80\% of the overall data are split, if federated, and used for training, while 20\% are held out for evaluation.

We instantiate the model with different choices for $g(x)$ -- with a linear predictor or with an NN, with and without PH (Table~\ref{table:models}). For baselines, we considered the works of \citet{Andreux2020FederatedModels} and \citet{Craig2021SurvivalProblem}. The former assumes PH, however, while the latter is a pooled data model. Both require upfront agreement on a specification of $f(t)$ to include non-PH, adding to the setup costs of a federation. We further note that no implementations of these models are available. We therefore provide the NN PH model to approximate the model of \citet{Andreux2020FederatedModels} -- a federated NN-based Cox model with PH -- and the Linear PH model as a standard baseline.

\begin{table}[htbp]
\centering
  \begin{tabular}{l c}
  Model & Predictor\\
  \toprule
  Linear PH & $\beta^\top x$\\
  NN PH & $g_{\phi}(x)$\\
  NN nonPH & $g_{\phi,t}(x)$\\
  \bottomrule
  \end{tabular}
\caption{Model choices.}
\label{table:models}
\end{table}

Architectures (Figure~\ref{fig:nn_detail}) are implemented in PyTorch 1.8.0 \citep{Paszke2019PyTorch:Library} with two hidden layers of 32 neurons for NN models and none for the linear model. Optimisation uses Adam with grid-sought learning rates ($10^{-1}$ to $10^{-5}$ on 20\% of the training data) and a batch size of 256. Base case discretisation uses 10 time steps.

\subsection{Results}
\label{sec:experiments:results}
In this section, we first compare the performance of the three models trained in a centralised fashion on pooled data against their federated performance on decentralised data. We next provide additional experiments exploring the impact of the discretisation grid chosen for the base case. We report averaged 5-fold cross-validation performance throughout.

\subsubsection{Federated Performance.}
On pooled data, the NN nonPH model outperforms or ties in concordance (Table~\ref{table:concordance}) and ties with the best in calibration (Table~\ref{table:brier}), indicating a gain from the relaxation of the PH assumption. For METABRIC and GBSG this aligns with \citet{Bellera2010VariablesCancer,Coradini2000Time-dependentCancer,Gore1984RegressionAuthor} who find non-PH amongst this patient group.

Comparing this to the federated setting with IID data, all three models maintain their performance (within one standard deviation) in concordance and calibration when aggregation is frequent. First hints of performance degradation amongst the NN-based models occur as aggregation becomes very infrequent (rightmost columns) while the Linear PH model appears largely unaffected. This observation is noteworthy, as infrequent aggregation will be a likely feature in practice given communication costs. This indicates a potential trade-off between model complexity and achievable aggregation frequency to support its training.

In practice, data are likely to be non-IID across centres. The results show that the performances of all three models suffer when this is the case. Generally, the NN-based models experience the most severe losses in performance and are largely outperformed by the Linear PH model. When aggregation is infrequent, the NN-based models on SUPPORT and GBSG effectively approach a no-skill predictor in concordance (average c-index of 0.5). Further, performance losses under infrequent aggregation of the NN-based models are, as would be expected, worse than under IID data. On SUPPORT, the NN-based models exhibit much greater performance differences than on the other datasets. In this respect, we note that SUPPORT has much longer survival times than METABRIC or GBSG (Table~\ref{table:data}), so that stratification by event time likely results in a more significantly different partition of the data for the former than for the latter two.

\begin{sidewaystable}[htbp]
\centering
    \small
    \begin{tabular}{c @{\hspace{1.25\tabcolsep}} c @{\hspace{1.25\tabcolsep}}  c c c | c c c | c c c}
    \addlinespace
    % \toprule
    &   &   \multicolumn{3}{c|}{METABRIC} & \multicolumn{3}{c|}{SUPPORT} & \multicolumn{3}{c}{GBSG} \\ 
    \cmidrule{3-5} \cmidrule{6-8} \cmidrule{9-11}
    &   &   \multicolumn{3}{c|}{Global / local rounds} & \multicolumn{3}{c|}{Global / local rounds} & \multicolumn{3}{c}{Global / local rounds}  \\ 
    \multirow{1}{*}{Data}  & \multirow{1}{*}{Model} & 100 / 1 & 20 / 5 & 1 / 100 & 100 / 1 & 20 / 5 & 1 / 100 & 100 / 1 & 20 / 5  & 1 / 100\\ \toprule
    \multirow{3}{*}{\rotatebox[origin=c]{90}{Pooled}}   & Linear PH & 63.5 $\!\pm\!$ 1.4 &               &               & 57.2 $\!\pm\!$ 1.0          &                      &                      & 66.5 $\!\pm\!$ 2.1 &             &            \\[0.25ex]
                                                        & NN PH     & 64.0 $\!\pm\!$ 0.6 &               &               & 60.8 $\!\pm\!$ 0.6          &                      &                      & 66.2 $\!\pm\!$ 2.6 &             &            \\[0.25ex]           
                                                        & NN nonPH  & \textbf{66.7 $\!\pm\!$ 2.1} &               &               & \textbf{61.5 $\!\pm\!$ 1.2} &                      &                      & \textbf{66.6 $\!\pm\!$ 1.9} &             &            \\[0.25ex]  \midrule
    \multirow{3}{*}{\rotatebox[origin=c]{90}{IID}}      & Linear PH & 63.9 $\!\pm\!$ 0.8 & 64.0 $\!\pm\!$ 2.2 & \textbf{63.7 $\!\pm\!$ 1.7} & 57.2 $\!\pm\!$ 0.8          & 57.2 $\!\pm\!$ 0.8          & 57.2 $\!\pm\!$ 0.4          & 66.5 $\!\pm\!$ 1.5 & 66.3 $\!\pm\!$ 0.6 & \textbf{66.3 $\!\pm\!$ 1.4} \\[0.25ex]
                                                        & NN PH     & 63.8 $\!\pm\!$ 1.6 & 62.5 $\!\pm\!$ 1.9 & 63.3 $\!\pm\!$ 1.2 & 60.6 $\!\pm\!$ 1.0          & 60.9 $\!\pm\!$ 0.9          & \textbf{58.3 $\!\pm\!$ 1.4} & \textbf{67.4 $\!\pm\!$ 1.7} & \textbf{67.1 $\!\pm\!$ 1.2} & 63.5 $\!\pm\!$ 2.6 \\[0.25ex]           
                                                        & NN nonPH  & \textbf{65.4 $\!\pm\!$ 1.9} & \textbf{65.7 $\!\pm\!$ 1.4} & 61.5 $\!\pm\!$ 2.1 & \textbf{62.1 $\!\pm\!$ 0.7} & \textbf{62.4 $\!\pm\!$ 0.4} & 56.7 $\!\pm\!$ 2.3          & 66.5 $\!\pm\!$ 1.0 & 66.4 $\!\pm\!$ 0.9 & 62.8 $\!\pm\!$ 1.9 \\[0.25ex] \midrule
    \multirow{3}{*}{\rotatebox[origin=c]{90}{Non-IID}}  & Linear PH & 59.2 $\!\pm\!$ 3.0 & \textbf{59.8 $\!\pm\!$ 2.1} & \textbf{61.0 $\!\pm\!$ 1.3} & 55.4 $\!\pm\!$ 1.5          & 56.1 $\!\pm\!$ 0.9          & \textbf{56.2 $\!\pm\!$ 0.5} & 61.1 $\!\pm\!$ 0.9 & \textbf{61.9 $\!\pm\!$ 3.0} & \textbf{56.5 $\!\pm\!$ 6.9} \\[0.25ex]
                                                        & NN PH     & \textbf{60.9 $\!\pm\!$ 1.1} & 59.4 $\!\pm\!$ 2.5 & 57.3 $\!\pm\!$ 4.2 & \textbf{57.0 $\!\pm\!$ 1.4} & \textbf{56.7 $\!\pm\!$ 1.1} & 52.9 $\!\pm\!$ 0.8          & \textbf{61.3 $\!\pm\!$ 4.3} & 61.8 $\!\pm\!$ 2.9 & 53.0 $\!\pm\!$ 6.2 \\[0.25ex]      
                                                        & NN nonPH  & 57.9 $\!\pm\!$ 2.9 & 59.6 $\!\pm\!$ 3.9 & 54.6 $\!\pm\!$ 5.3 & 50.8 $\!\pm\!$ 0.6          & 50.9 $\!\pm\!$ 0.9          & 50.8 $\!\pm\!$ 1.2          & 57.6 $\!\pm\!$ 1.9 & 55.8 $\!\pm\!$ 2.0 & 52.0 $\!\pm\!$ 2.5 \\[0.25ex] \bottomrule
    \end{tabular}
\caption{C-index (rebased to 100) -- mean and standard deviation. Higher values are better.}
\label{table:concordance}
% \vspace{10pt}
\centering
\small
    \begin{tabular}{c @{\hspace{1.25\tabcolsep}} c @{\hspace{1.25\tabcolsep}}  c c c | c c c | c c c}
    \addlinespace
    % \toprule
    &   &   \multicolumn{3}{c|}{METABRIC} & \multicolumn{3}{c|}{SUPPORT} & \multicolumn{3}{c}{GBSG} \\ 
    \cmidrule{3-5} \cmidrule{6-8} \cmidrule{9-11}
    &   &   \multicolumn{3}{c|}{Global / local rounds} & \multicolumn{3}{c|}{Global / local rounds} & \multicolumn{3}{c}{Global / local rounds}  \\ 
    \multirow{1}{*}{Data}  & \multirow{1}{*}{Model} & 100 / 1 & 20 / 5 & 1 / 100 & 100 / 1 & 20 / 5 & 1 / 100 & 100 / 1 & 20 / 5  & 1 / 100\\ \toprule
    \multirow{3}{*}{\rotatebox[origin=c]{90}{Pooled}}   & Linear PH & \textbf{16.4 $\!\pm\!$   0.6} &                      &                      & 20.9 $\!\pm\!$ 0.5          &                      &                      & \textbf{18.0 $\!\pm\!$ 0.5}          &                      &                      \\[0.25ex]
                                                        & NN PH     & 16.8 $\!\pm\!$ 0.7            &                      &                      & \textbf{19.6 $\!\pm\!$ 0.4}          &                      &                      & 18.2 $\!\pm\!$ 0.8          &                      &                      \\[0.25ex]           
                                                        & NN nonPH  & \textbf{16.4 $\!\pm\!$ 0.8}            &                      &                      & \textbf{19.6 $\!\pm\!$ 0.4} &                      &                      & \textbf{18.0 $\!\pm\!$ 0.4} &                      &                      \\[0.25ex]  \midrule
    \multirow{3}{*}{\rotatebox[origin=c]{90}{IID}}      & Linear PH & \textbf{16.3 $\!\pm\!$ 1.1}   & 16.7 $\!\pm\!$ 0.9          & \textbf{16.5 $\!\pm\!$ 0.7} & 20.9 $\!\pm\!$ 0.6          & 20.9 $\!\pm\!$ 0.3          & \textbf{20.9 $\!\pm\!$ 0.6} & 18.1 $\!\pm\!$ 0.2          & 18.1 $\!\pm\!$ 0.2          & \textbf{18.1 $\!\pm\!$ 0.3} \\[0.25ex]
                                                        & NN PH     & 17.2 $\!\pm\!$ 1.1            & 18.1 $\!\pm\!$ 0.5          & 18.1 $\!\pm\!$ 1.1          & 19.7 $\!\pm\!$ 0.7          & 19.8 $\!\pm\!$ 0.2          & 22.6 $\!\pm\!$ 1.3          & \textbf{17.7 $\!\pm\!$ 0.3} & \textbf{17.8 $\!\pm\!$ 0.4} & 19.8 $\!\pm\!$ 1.9          \\[0.25ex]           
                                                        & NN nonPH  & \textbf{16.3 $\!\pm\!$ 1.3}            & \textbf{16.5 $\!\pm\!$ 0.6} & 19.1 $\!\pm\!$ 0.8          & \textbf{19.5 $\!\pm\!$ 0.7} & \textbf{19.4 $\!\pm\!$ 0.4} & 21.1 $\!\pm\!$ 0.6          & 18.3 $\!\pm\!$ 0.5          & 18.2 $\!\pm\!$ 0.6          & 21.7 $\!\pm\!$ 1.5          \\[0.25ex] \midrule
    \multirow{3}{*}{\rotatebox[origin=c]{90}{Non-IID}}  & Linear PH & \textbf{18.3 $\!\pm\!$ 0.8}            & \textbf{18.0 $\!\pm\!$ 0.6} & \textbf{19.3 $\!\pm\!$ 0.7} & 22.9 $\!\pm\!$ 0.7          & 22.1 $\!\pm\!$ 0.3          & \textbf{21.4 $\!\pm\!$ 0.3} & \textbf{20.4 $\!\pm\!$ 0.5} & \textbf{20.1 $\!\pm\!$ 0.5} & \textbf{22.1 $\!\pm\!$ 0.3} \\[0.25ex]
                                                        & NN PH     & \textbf{18.3 $\!\pm\!$ 0.5}   & 18.1 $\!\pm\!$ 0.2          & 19.7 $\!\pm\!$ 1.0          & \textbf{22.6 $\!\pm\!$ 0.5} & \textbf{21.9 $\!\pm\!$ 0.2} & 24.0 $\!\pm\!$ 1.8          & 20.5 $\!\pm\!$ 0.6          & 20.6 $\!\pm\!$ 0.4          & 22.3 $\!\pm\!$ 0.5          \\[0.25ex]      
                                                        & NN nonPH  & 20.9 $\!\pm\!$ 0.3            & 20.1 $\!\pm\!$ 0.5           & 21.1 $\!\pm\!$ 1.0          & 26.2 $\!\pm\!$ 0.4          & 25.3 $\!\pm\!$ 0.5          & 25.3 $\!\pm\!$ 0.7          & 23.6 $\!\pm\!$ 0.9          & 22.4 $\!\pm\!$ 0.8          & 22.7 $\!\pm\!$ 1.0          \\[0.25ex] \bottomrule
    \end{tabular}
\caption{Integrated Brier scores (rebased to 100) -- mean and standard deviation. Lower values are better.}
\label{table:brier}
\end{sidewaystable}

\subsubsection{Impact of Discretisation Fineness.}

\begin{figure*}[htbp]
    \centering
    \begin{subfigure}[b]{\linewidth}
        \centering
        \includegraphics[width=\linewidth]{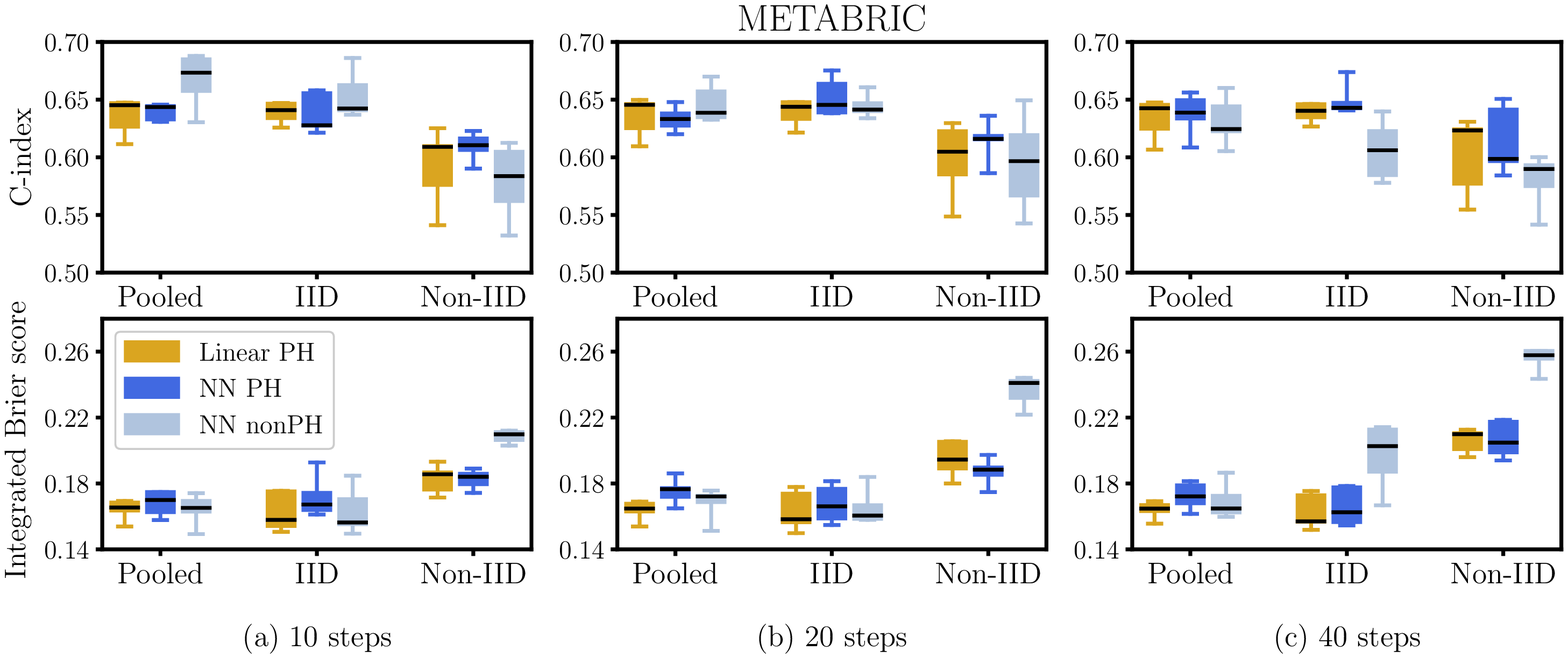}%
    \end{subfigure}
    % \hfill
    \begin{subfigure}[b]{\linewidth}
        \centering
        \includegraphics[width=\linewidth]{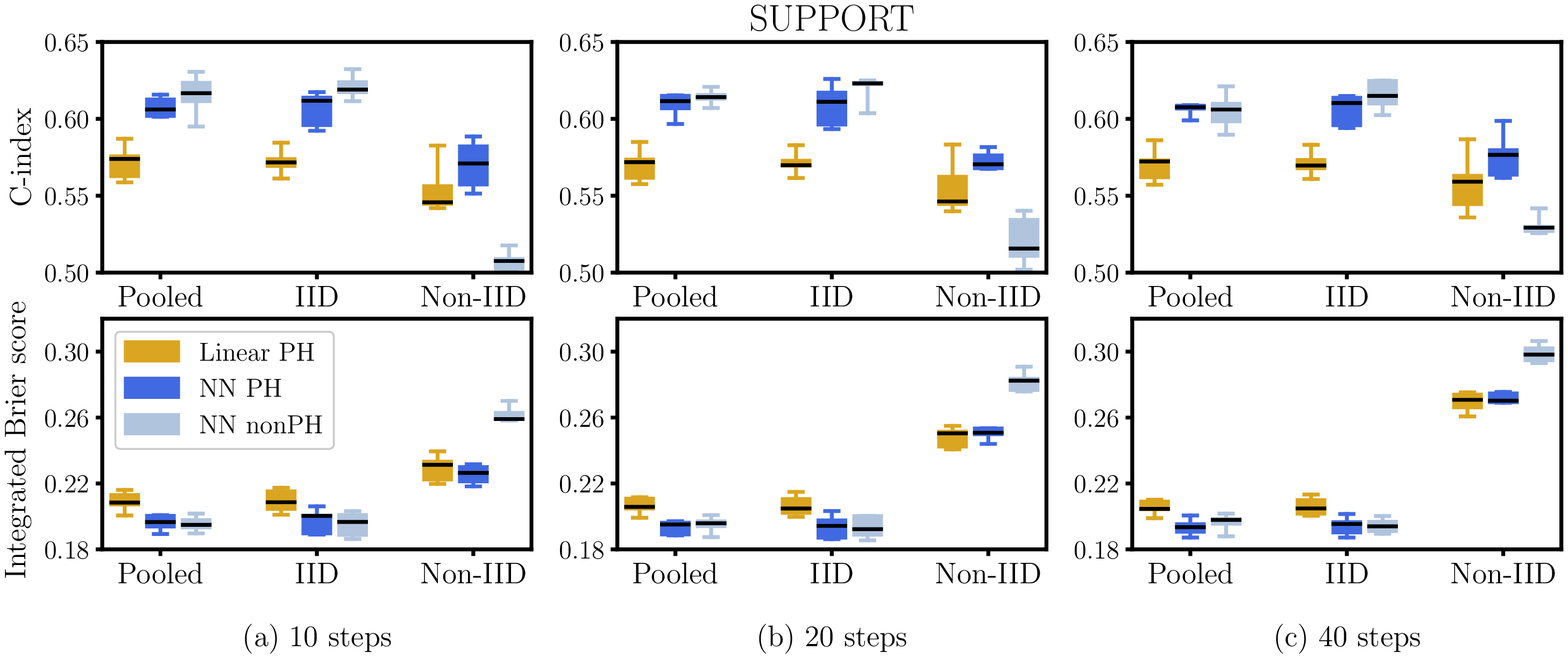}
    \end{subfigure}
    % \hfill
    \begin{subfigure}[b]{\linewidth}
        \centering
        \includegraphics[width=\linewidth]{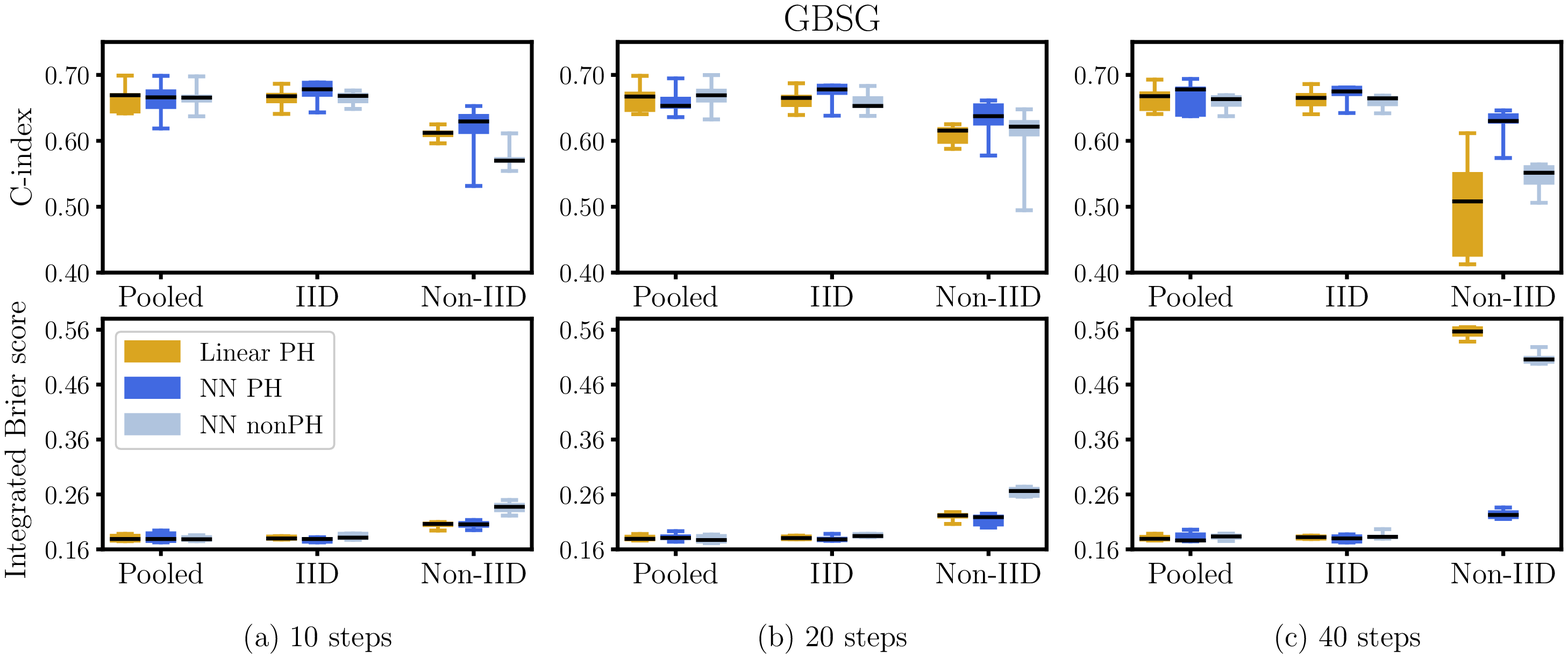}
    \end{subfigure}    
    \caption{Model performance with increasing discretisation fineness. Federated models were trained with 100 global and 1 local rounds. Performance decreases on smaller METABRIC and GBSG datasets with mixed results on the larger SUPPORT dataset.}
    \label{fig:discretisation_finess}
\end{figure*}

We re-train models on finer time grids using 100 global and 1 local rounds. A finer grid on METABRIC (Figure~\ref{fig:discretisation_finess} upper panel) and GBSG (Figure~\ref{fig:discretisation_finess} lower panel) did not improve performance and, in fact, appears to degrade performance. Notably, the Linear PH model becomes a no-skill predictor in terms of concordance on the non-IID GBSG case. The results are less conclusive on SUPPORT (Figure~\ref{fig:discretisation_finess} middle panel), as a finer time grid appears to result in a minor to no increase in concordance at the expense of a loss in calibration. 

A finer time grid can be expected to result in a trade-off between closer approximation of true (smooth) survival and a reduction in data available in any given time step. An increase from 10 to 20 time steps, for instance, halves the number of available data points to estimate a given step. The latter effect appears to dominate on the smaller METABRIC and GBSG datasets, and less so for the approximately 4-times larger SUPPORT dataset.  

% \begin{figure*}[htbp]
%     \centering
%     \includegraphics[width=\linewidth]{figs/duration_flex_metabric.eps}
%     \caption{Model performance with finer time grids. Concordance and calibration worsen with increasing fineness.}
%     \label{fig:discretisation_finess_metabric}
% \end{figure*}

% \begin{figure*}[hbtp]
%     \centering
%     \includegraphics[width=\linewidth]{figs/duration_flex_support.eps}
%     \caption{Model performance with finer time grids. A mostly stable c-index accompanies a worsening in calibration with increasing fineness.}
%     \label{fig:discretisation_finess_support}
% \end{figure*}

% \begin{figure*}[hbtp]
%     \centering
%     \includegraphics[width=\linewidth]{figs/duration_flex_gbsg.eps}
%     \caption{Model performance with finer time grids. Concordance and calibration worsen with increasing fineness.}
%     \label{fig:discretisation_finess_gbsg}
% \end{figure*}

%  \citep{Li2018FederatedNetworks} note may result in performance losses on non-IID data, as is evident in columns three and four of Table~\ref{table:concordance}.

% The Brier scores (Table~\ref{table:brier}) show that all three models are equally well calibrated on pooled and IID data with frequent aggregation but degrade in the non-IID case or under sparse aggregation. Notably, while the linear model shows good concordance throughout, its calibration worsens on non-IID data, pointing to the importance of including a calibration measure when evaluating model performance.

%% file: 06discussion.tex
\section{Conclusion}
\label{sec:discussion}
We present a federated Cox model that relaxes the proportional hazards (PH) assumption and demonstrate its ability to maintain concordance and calibration relative to a pooled baseline under various linearity and PH assumptions. Compared to prior work, this federation scheme encodes the decision between PH and non-PH in a binary choice over the output layer, rather than requiring upfront agreement on a specification of $f(t)$. We note that our model is not restricted to a particular data type or network architecture excepting the output component. Future work could adapt the model for image-based federated survival predictions. 

The decrease in performance on non-IID data (even if pathologically derived in this paper) represents a challenge to the application of federated learning in practice. Extensions could include exploring methods accounting for statistical heterogeneity \citep{Yang2020HeterogeneousLearning,Li2018FederatedNetworks} or other federation topologies which maintain locally specialised models trained in a peer-to-peer fashion \citep{Rieke2020}. While the heterogeneity in this paper was derived from label stratification, other types of heterogeneity, such as covariate shifts, could be explored: for image-based survival predictions, differences in acquisition protocols could provide one such avenue.

%% file: appxKM.tex
\section{Additional Figures}
\label{sec:appx}

% \subsection{Discretisation and Interpolation}
\begin{figure}[hbtp]
\begin{subfigure}[h]{0.49\linewidth}
    \centering
    \includegraphics[width=\linewidth]{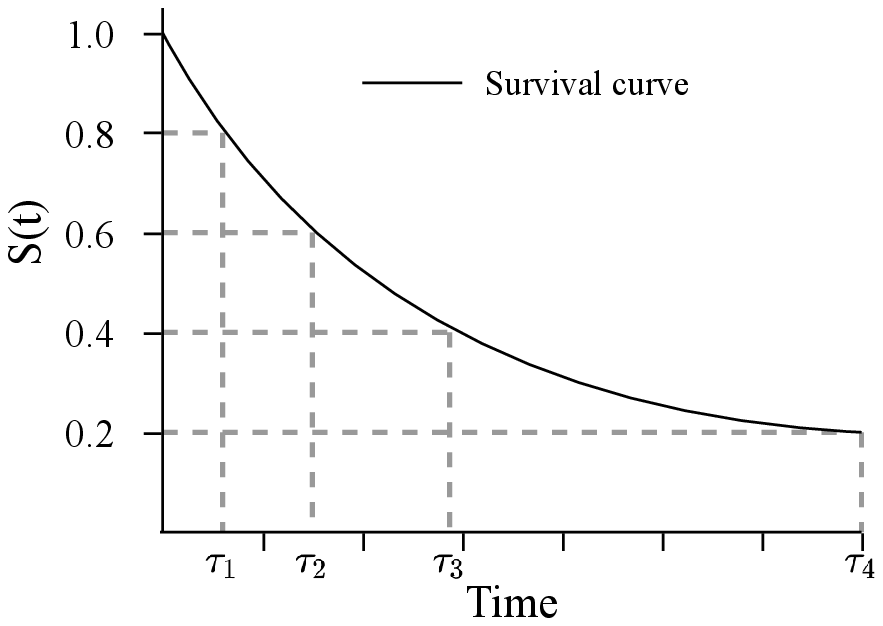}
    \caption{Kaplan-Meier-based discretisation.}
    \label{fig:km-discretisation}
\end{subfigure}
\begin{subfigure}[h]{0.49\linewidth}
    \centering
    \includegraphics[width=\linewidth]{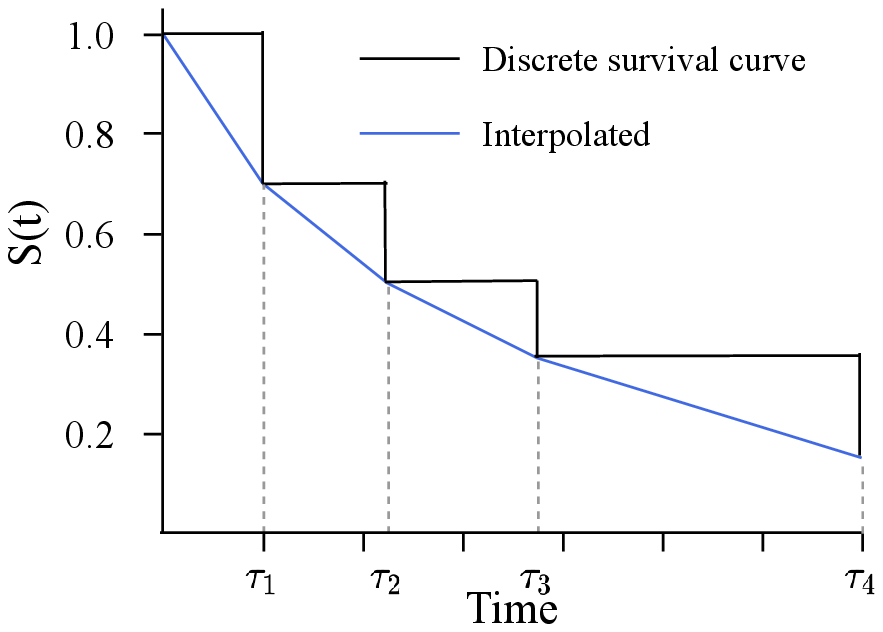}
    \caption{Constant density interpolation.}
    \label{fig:interpolation-cdi}
\end{subfigure}
\caption{Discretisation and interpolation.}
\end{figure}

% \begin{figure}[ht]
%     \centering
%     \includegraphics[width=0.7\linewidth]{figs/KM_discretisation.png}
%     \caption{Discretisation based on Kaplan-Meier quantiles with 4 time steps.}
%     \label{fig:km-discretisation}
% \end{figure}

% \begin{figure}[hb]
%     \centering
%     \includegraphics[width=0.7\linewidth]{figs/interpolation.png}
%     \caption{A discrete survival curve interpolated using constant density interpolation.}
%     \label{fig:interpolation-cdi}
% \end{figure}

% \subsection{Calibration}
\begin{figure}[h]
    \centering
    \includegraphics[width=0.7\linewidth]{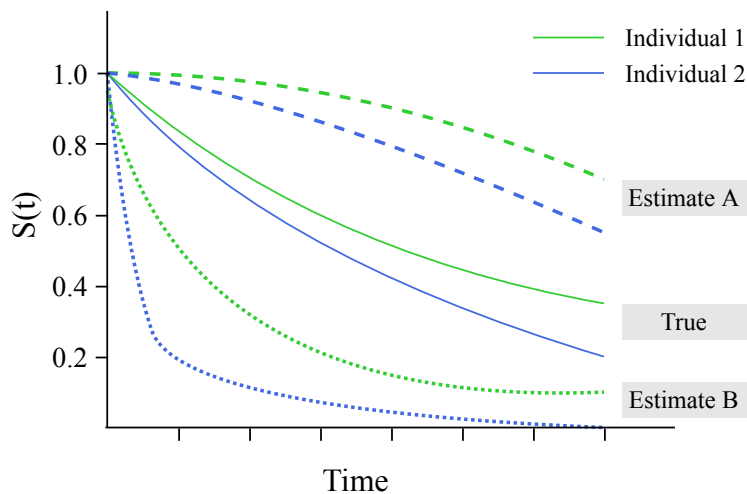}
    \caption{Two sets of survival estimates with correct ranking (green above blue) but poor calibration given under- / overestimation of true survival curves.}
    \label{fig:calibration}
\end{figure}

% \newpage
% \subsection{Kaplan-Meier Estimates}
\begin{figure}[h]
    \centering
    \includegraphics[width=\linewidth]{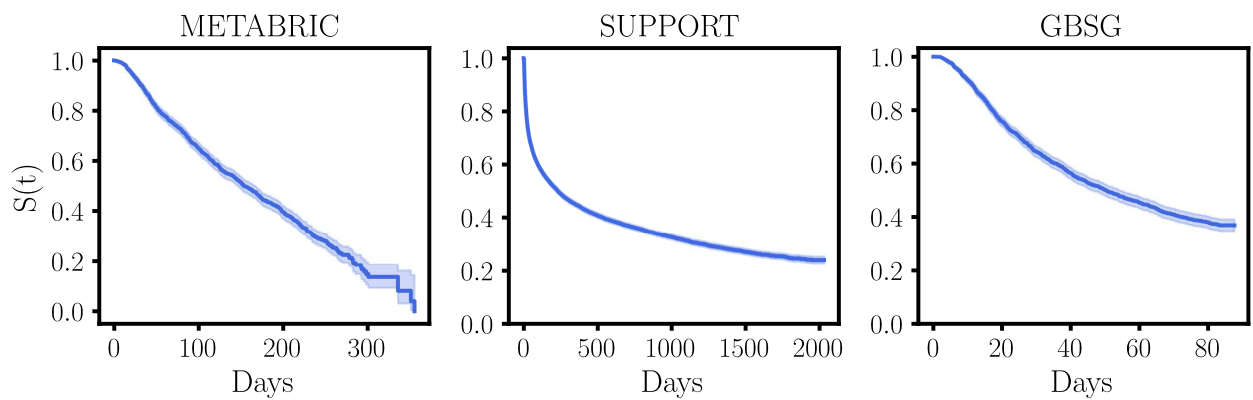}
    \caption{Kaplan-Meier estimates with 95\% confidence interval.}
    \label{fig:km}
\end{figure}

% \begin{figure}[htbp]
%     \centering
%     \begin{subfigure}[t]{0.7\linewidth}
%         \includegraphics[width=\linewidth]{figs/KM-metabric.png}%
%     \end{subfigure}
%     % \hfill
%     \begin{subfigure}[t]{0.7\linewidth}
%         \includegraphics[width=\linewidth]{figs/KM-support.png}
%     \end{subfigure}
%     % \hfill
%     \begin{subfigure}[t]{0.7\linewidth}
%         \includegraphics[width=\linewidth]{figs/KM-gbsg.png}
%     \end{subfigure}    
%     \caption{Kaplan-Meier estimates with 95\% confidence interval.}
%     \label{fig:km}
% \end{figure}